\pdfoutput=1

\documentclass[11pt]{article}

\usepackage[]{EMNLP2022}

\usepackage{times}
\usepackage{latexsym}

\usepackage[T1]{fontenc}

\usepackage[utf8]{inputenc}

\usepackage{microtype}

\usepackage{inconsolata}

\usepackage{amsmath}
\DeclareMathOperator{\selfattn}{SelfAttn}
\DeclareMathOperator{\sa}{S.A.}
\DeclareMathOperator{\softmax}{softmax}

\DeclareMathOperator{\syn}{Syntax}
\DeclareMathOperator{\sy}{Syn.}
\DeclareMathOperator{\comp}{Composition}
\DeclareMathOperator{\com}{C.}

\usepackage{lingmacros}
\newlength{\astspace}
\newcommand{\un}{*}

\newcommand{\ac}{\hspace*{\astspace}}

\usepackage{booktabs,multirow}
\usepackage{graphicx}
\usepackage{colortbl}
\usepackage{nicematrix}
\usepackage{qtree}

\usepackage{caption}
\usepackage[subrefformat=parens]{subcaption}
\usepackage{comment}
\usepackage{hyperref}
\hypersetup{
    breaklinks=true,   
    colorlinks=true,   
}
\title{Composition, Attention, or Both?\thanks{~~While writing this paper, we noticed that \citet{https://doi.org/10.48550/arxiv.2203.00633} was submitted to the arXiv, proposing Transformer Grammars (TGs) that incorporate recursive syntactic composition via an attention mask. Their work and ours are similar in spirit, but different in how to obtain a vector representation of subtrees, making them complementary. We discuss the details in Section~\ref{sec:releted}.}}

\author{Ryo Yoshida \and Yohei Oseki \\
  The University of Tokyo \\
  \texttt{\{yoshiryo0617, oseki\}@g.ecc.u-tokyo.ac.jp}}

\begin{document}
\maketitle
\begin{abstract}
In this paper, we propose a novel architecture called \textbf{Composition Attention Grammars} (CAGs) that recursively compose subtrees into a single vector representation with a composition function, and selectively attend to previous structural information with a self-attention mechanism. We investigate whether these components---the composition function and the self-attention mechanism---can both induce human-like syntactic generalization. Specifically, we train language models (LMs) with and without these two components with the model sizes carefully controlled, and evaluate their syntactic generalization performance against six test circuits on the SyntaxGym benchmark. The results demonstrated that the composition function and the self-attention mechanism both play an important role to make LMs more human-like, and closer inspection of grammatical phenomena implied that the composition function allowed syntactic features, but not semantic features, to percolate into subtree representations.

\end{abstract}
\begin{figure*}
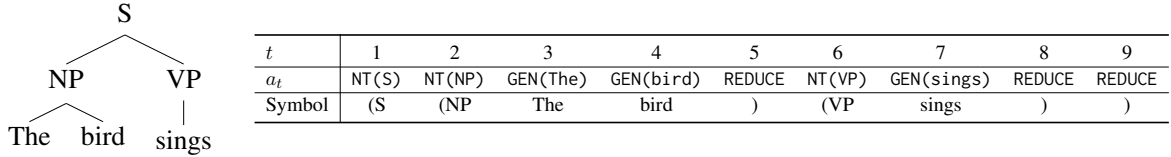

    \centering
    \scalebox{0.99}{
    \begin{tabular}{cc}
        \begin{minipage}{0.19\hsize}
            \begin{flushleft}
            \scalebox{0.95}{
            \Tree[.S
                    [.NP
                        {The}
                        {bird}
                    ]
                    [.VP
                        {sings}
                    ]
                ]
            }
            \end{flushleft}
        \end{minipage}%
        %
        %
        \begin{minipage}{0.78\hsize}
            \begin{flushright}
            \scalebox{0.70}{
            \begin{NiceTabular}{l|ccccccccc}
             \toprule
                $t$ & 1 & 2 & 3 & 4 & 5 & 6 & 7 & 8 & 9 \\
                \hline
                $a_t$ & \texttt{NT(S)} & \texttt{NT(NP)} & \texttt{GEN(The)} & \texttt{GEN(bird)} & \texttt{REDUCE} & \texttt{NT(VP)} & \texttt{GEN(sings)} & \texttt{REDUCE} & \texttt{REDUCE}\\
                \hline
                Symbol & (S & (NP & The & bird & ) & (VP & sings & ) & ) \\
            \bottomrule
            \end{NiceTabular}}
            \end{flushright}
        \end{minipage}
    \end{tabular}
    }
    \caption{An example of actions to jointly generate the sentence and its syntactic structure in a top-down, left-to-right fashion.}
    \label{fig:action}
\end{figure*}

\section{Introduction}
\label{sec:intro}
Recently, language models (LMs) trained on large datasets have achieved remarkable success in various Natural Language Processing (NLP) tasks (cf. \citealp{1905.00537, wang2018glue}). The literature of targeted syntactic evaluations has shown that these models implicitly learn syntactic structures of natural language, even though they do not receive explicit syntactic supervision \citep{warstadt-etal-2020-blimp-benchmark, hu-etal-2020-systematic}.

However, previous work has also shown that there is still a benefit for LMs to receive explicit syntactic supervision. Recurrent Neural Network Grammars (RNNGs; \citealp{dyer-etal-2016-recurrent}), the integration of Recurrent Neural Networks (RNNs; \citealp{Elman:1990}) with an explicit syntactic bias, have achieved better syntactic generalization performance than vanilla RNNs \citep{kuncoro-etal-2018-lstms, wilcox-etal-2019-structural, hu-etal-2020-systematic}. In addition, previous work has recommended RNNGs as a cognitively plausible architecture, showing that RNNGs can successfully predict human reading times \citep{yoshida-etal-2021-modeling} or brain activities \citep{hale-etal-2018-finding}. The key difference between RNNGs and RNNs is a \textbf{composition function}, which recursively composes subtrees into a single vector representation. 

On the other hand, Transformer architectures \citep{Vaswani2017AttentionNeed} have been shown to outperform RNN architectures in various NLP tasks \citep{devlin-etal-2019-bert}. The key difference between Transformers and RNNs here is a \textbf{self-attention mechanism}, which selectively attends to previous vectors to obtain sentence representations. Recently, an attempt was made to investigate whether Transformer architectures with the self-attention mechanism also benefit from explicit syntactic supervision \citep{qian-etal-2021-structural}, but their ``Parsing as Language Modeling (PLM)'' approach \citep{choe-charniak-2016-parsing} does not employ the composition function, which is essential for RNNGs. Therefore, it is reasonable to hypothesize that their approach may not achieve the full benefit of explicit syntactic supervision.

In this paper, we propose a novel architecture called \textbf{Composition Attention Grammars} (CAGs) that recursively compose subtrees into a single vector representation with the composition function, and selectively attend to previous structural information with the self-attention mechanism. We investigate whether these components---the composition function and the self-attention mechanism---can both induce human-like syntactic generalization. Specifically, we train LMs with and without these two components, with the model sizes carefully controlled, and evaluate their syntactic generalization performance against six test circuits \citep{hu-etal-2020-systematic} on the SyntaxGym benchmark \citep{gauthier-etal-2020-syntaxgym}. The results demonstrated that the composition function and the self-attention mechanism both play an important role to make LMs more human-like, and closer inspection of grammatical phenomena implied that the composition function allowed syntactic features, but not semantic features, to percolate into subtree representations.

In addition, the methodological innovation of this paper is a strictly controlled experimental design, as practiced in cognitive sciences. In NLP research, evaluations are often conducted on models with different model sizes, leading to uncertainty regarding which component of these models affects the results. This paper conducts strictly controlled experiments in order to isolate the effects of individual components such as the composition function and the self-attention mechanism.

\section{Composition Attention Grammar}
\label{sec:CAG}
In this section, we introduce a novel architecture called Composition Attention Grammars (CAGs).
\begin{figure*}[t]
    \centering
    \includegraphics[width=\hsize]{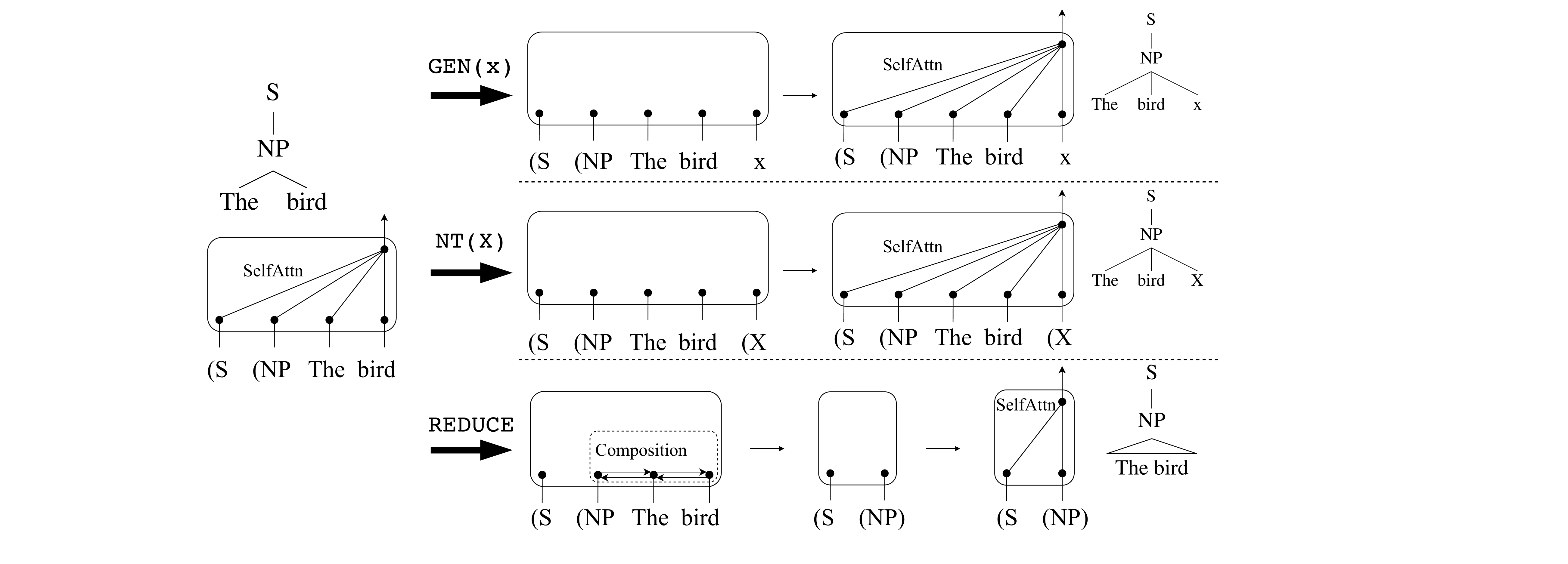}
    \caption{The architecture of Composition Attention Grammars (CAGs). CAGs utilize (i) the composition function to recursively compose subtrees into a single vector representation, and (ii) the self-attention mechanism to selectively attend to previous structural information.}
    \label{fig:CAG}
\end{figure*}

\subsection{Syntactic language model}
\label{subsec:actions}
CAGs are a type of syntactic LM \citep{choe-charniak-2016-parsing, dyer-etal-2016-recurrent, qian-etal-2021-structural}, which estimates the following joint distribution of a sentence $X$ and its syntactic structure $Y$:
\begin{align}
\label{eq: joint_dist}
    p(X,Y) = p(a_1, \cdots, a_n) = \prod_{t=1}^{n}p(a_t|a_{<t})
\end{align}
where $a_t$ is an action by which CAGs jointly generate the sentence and its syntactic structure in a top-down, left-to-right fashion. Each $a_t$ can be one of the three actions below:
\begin{itemize}
    \item \texttt{GEN(x)}: Generate a terminal symbol  ``x''.
    \item \texttt{NT(X)}: Open a nonterminal symbol ``X''.
    \item \texttt{REDUCE}: Close a nonterminal symbol that was opened by \texttt{NT(X)}.
\end{itemize}
See Figure~\ref{fig:action} for an example of actions to jointly generate the sentence and its syntactic structure in a top-down, left-to-right fashion.

\subsection{Architecture}
To estimate the joint distribution in Equation~\ref{eq: joint_dist}, CAGs utilize (i) the composition function to recursively compose subtrees into a single vector representation, and (ii) the self-attention mechanism to selectively attend to previous structural information. The architecture of CAGs is summarized in Figure~\ref{fig:CAG}. Following previous work \citep{kuncoro-etal-2017-recurrent, noji-oseki-2021-effective}, CAGs rely on a stack data structure, and each action in Section~\ref{subsec:actions} changes the stack state as follows:
\begin{itemize}
    \item \texttt{GEN(x)}: Push a terminal embedding $\mathbf{e}_x$ onto the stack.
    \item \texttt{NT(X)}: Push a nonterminal embedding $\mathbf{e}_X$ onto the stack.
    \item \texttt{REDUCE}: First, repeatedly pop vectors from the stack until a nonterminal embedding is popped. Then, apply the composition function based on bidirectional LSTMs \citep{650093} to these popped vectors $\mathbf{e}_l, \dots, \mathbf{e}_m$, to compose subtrees into a single vector representation $\mathbf{e}_s$:
    \begin{align}
    \mathbf{e}_s = \comp([\mathbf{e}_l, \dots, \mathbf{e}_m]).
\end{align}
    $\mathbf{e}_s$ is then pushed onto the stack.
\end{itemize}

After each action, CAGs employ the self-attention mechanism, which selectively attends to previous vectors in the stack $\mathbf{e}_1, \dots, \mathbf{e}_k$ by calculating the weight of attention to each vector with the query, key, and value vectors generated from $\mathbf{e}_1, \dots, \mathbf{e}_k$, in order to represent a partial parse at each time step $t$:
\begin{align}
    \mathbf{h}_t = \selfattn([\mathbf{e}_1, \dots, \mathbf{e}_k]).
\end{align}
Then, $\mathbf{h}_t$ defines the next action distribution:
\begin{align}
    a_{t+1} \sim \softmax(\mathbf{W}_a\mathbf{h}_t+\mathbf{b}_a)
\end{align}
where $\mathbf{W}_a$ and $\mathbf{b}_a$ are the weights and biases of a fully connected layer that projects $\mathbf{h}_t$ to logits for each action $a$, and $\softmax$ is a softmax function that projects the logits to the next action distribution.

\subsection{Differences from other syntactic LMs}
In this subsection, we focus on the differences between CAGs and other syntactic LMs.
\paragraph{Difference from RNNGs} 
CAGs and RNNGs both utilize the composition function to recursively compose subtrees into a single vector representation. CAGs differ from RNNGs in that, in order to represent the partial parse at each time step, CAGs utilize the self-attention mechanism which selectively attends to previous structural information, whereas RNNGs utilize stack-LSTMs \cite{dyer-etal-2015-transition}. We hypothesize that CAGs have the advantage of selective attention to previous structural information over RNNGs.

\paragraph{Difference from PLMs}
CAGs and PLMs both utilize the self-attention mechanism which selectively attends to previous structural information. CAGs differ from PLMs in that CAGs utilize the composition function to recursively compose subtrees into a single vector representation, whereas PLMs treat actions $a_1, \dots, a_n$ flatly as vanilla Transformers treat words $w_1, ..., w_n$. We hypothesize that CAGs have the advantage of recursive composition of subtrees over PLMs.

In order to incorporate composition-like characteristics, \citet{qian-etal-2021-structural} proposed PLM-masks, namely, PLMs with a dynamic masking mechanism, which specializes two attention heads: one to attend to the inside of the most recently opened nonterminal symbol, and another to attend to the outside. We will perform a comparison between CAGs and PLM-masks in order to investigate whether recursive composition of subtrees has additional advantages over the dynamic masking mechanism in inducing human-like syntactic generalization.

\begin{table*}[t]
    \centering
    \scalebox{0.77}{
    \begin{NiceTabular}{l||c|c|c}
        \toprule
        \multicolumn{1}{c}{}&\multicolumn{1}{c}{$-\syn$}&\multicolumn{2}{c}{$+\syn$}\\
        \cmidrule{3-4}
        \multicolumn{1}{c}{}&\multicolumn{1}{c}{}&\multicolumn{1}{c}{$-\comp$}&\multicolumn{1}{c}{$+\comp$}\\
        \hline\hline
        $-\selfattn$ & \begin{tabular}{c}LSTM\\ \citep{hochreiterLongShorttermMemory1997}\end{tabular} & \begin{tabular}{c}ActionLSTM\\ \citep{choe-charniak-2016-parsing}\end{tabular} & \begin{tabular}{c}RNNG\\ \citep{dyer-etal-2016-recurrent}\end{tabular} \\
        \hline
        $+\selfattn$ & \begin{tabular}{c}Transformer\\ \citep{radford2018improving}\end{tabular} & \begin{tabular}{c}PLM\\ \citep{qian-etal-2021-structural}\end{tabular} & \begin{tabular}{c}PLM-mask\\($(+)\comp$; \citealp{qian-etal-2021-structural})\\  \rowcolor[rgb]{0.83, 0.83, 0.83} CAG\\ (This work) \end{tabular} \\
        \bottomrule 
    \end{NiceTabular}}
    \caption{LMs investigated in this paper. $\pm\syn$ means whether LMs receive explicit syntactic supervision. $\pm\comp$ means whether LMs utilize the composition function, and $\pm\selfattn$ means whether LMs are based on Transformer architectures with the self-attention mechanism. PLM-masks do not utilize the composition function, but use the local subtree information with the dynamic masking mechanism ($(+)\comp$).}
    \label{tab:models}
\end{table*}

\begin{table*}[t]
    \centering
    \scalebox{0.90}{
    \begin{tabular}{lccccc}
        \toprule
          & \#Layer & \#Hidden dimension & \#Input dimension & \#Head & \#Model size\\
        \hline
        LSTM & 2 & 301 & 301 & N/A & \textbf{16.59M} \\
        ActionLSTM & 2 & 301 & 301 & N/A & \textbf{16.58M} \\
        RNNG & 2 & 276 & 276 & N/A & \textbf{16.61M} \\
        Transformer & 3 & 272 & 272 & 4 & \textbf{16.62M} \\
        PLM & 3 & 272 & 272 & 4 & \textbf{16.63M} \\
        PLM-mask & 3 & 272 & 272 & 4 & \textbf{16.63M} \\
        CAG & 3 & 256 & 256 & 4 & \textbf{16.57M} \\
        \bottomrule
    \end{tabular}}
    \caption{Hyperparameters of LMs investigated in this paper. We controlled the hyperparameters in order to make model sizes maximally comparable.}
    \label{tab:model_size}
\end{table*}

\section{Experiment}
We designed a strictly controlled experiment for testing whether the two components---the composition function and the self-attention mechanism---can both induce human-like syntactic generalization. Specifically, we train LMs with and without these two components with the model sizes carefully controlled, and evaluate their syntactic generalization performance against six test circuits on the SyntaxGym benchmark. We also train and evaluate two vanilla LMs with and without the self-attention mechanisms as a baseline. The following subsections describe the experimental settings in further detail.

\subsection{Language models}
\label{subsec:LM}
This subsection describes LMs investigated in this paper (Table~\ref{tab:models}). We controlled the hyperparameters in order to make model sizes maximally comparable (Table~\ref{tab:model_size}).

\paragraph{LSTM}
LSTMs \citep{hochreiterLongShorttermMemory1997} are a vanilla LM ($-\syn$) based on RNN architectures ($-\selfattn$). LSTMs were adopted as a baseline for syntactic LMs without the self-attention mechanism. Our LSTMs were implemented with the PyTorch package.\footnote{\url{https://github.com/pytorch/pytorch}}

\paragraph{ActionLSTM}
ActionLSTMs \citep{choe-charniak-2016-parsing} are a syntactic LM ($+\syn$) based on RNN architectures ($-\selfattn$). ActionLSTMs treat actions flatly without the composition function ($-\comp$). Our ActionLSTMs were implemented with the PyTorch package.

\paragraph{RNNG}
RNNGs are a syntactic LM ($+\syn$) based on RNN architectures ($-\selfattn$). RNNGs recursively compose subtrees into a single vector representation with the composition function ($+\comp$). The implementation with the PyTorch package by \citet{noji-oseki-2021-effective} was employed.\footnote{\url{https://github.com/aistairc/rnng-pytorch}}

\paragraph{Transformer}
Transformers \citep{radford2018improving} are a vanilla LM ($-\syn$) based on Transformer architectures ($+\selfattn$). Transformers were adopted as a baseline for syntactic LMs with the self-attention mechanism. Our Transformers were implemented with Huggingface's Transformer package \citep{wolf-etal-2020-transformers}.\footnote{\url{https://github.com/huggingface/transformers}}

\paragraph{PLM}
PLMs are a syntactic LM ($+\syn$) based on Transformer architectures ($+\selfattn$). PLMs treat actions flatly without the composition function ($-\comp$). The implementation with Huggingface's Transformer package by \citet{qian-etal-2021-structural} was employed.\footnote{\url{https://github.com/IBM/transformers-struct-guidance}}

\paragraph{PLM-mask}
PLM-masks are a syntactic LM ($+\syn$) based on Transformer architectures ($+\selfattn$). PLM-masks do not utilize the composition function, but use the local subtree information with the dynamic masking mechanism ($(+)\comp$). The implementation with Huggingface’s Transformer package by \citet{qian-etal-2021-structural} was employed.

\paragraph{CAG}
CAGs are a syntactic LM ($+\syn$) based on Transformer architectures ($+\selfattn$). CAGs recursively compose subtrees into a single vector representation with the composition function ($+\comp$). Our CAGs were implemented with the PyTorch and Huggingface's Transformer packages.\footnote{Our implementation is available at \url{https://github.com/osekilab/CAG}. The implementation is based on the PyTorch implementation of RNNG by \citet{noji-oseki-2021-effective}.}

\subsection{Training}
All LMs were trained on the BLLIP-\textsc{lg} dataset, which comprises 1.8M sentences and 42M tokens sampled from the Brown Laboratory for Linguistic Information Processing 1987-89 Corpus Release 1 (BLLIP; \citealp{charniak2000bllip}). We followed the train-dev-test split of \citet{hu-etal-2020-systematic}. Following \citet{qian-etal-2021-structural}, we split the sentences into subwords using a Byte Pair Encoding tokenizer \citep{sennrich-etal-2016-neural} from Huggingface's Transformer package. The baseline vanilla LMs used only terminal subwords, whereas the syntactic LMs used terminal subwords and syntactic structures. We utilized syntactic structures re-parsed by \citet{hu-etal-2020-systematic} with a state-of-the-art constituency parser \citep{kitaev-klein-2018-constituency}. All LMs were trained at the sentence level with a learning rate of $10^{-3}$, a dropout rate of $0.1$, Adam optimizer, and a minibatch size of 256 for 15 epochs. We selected the checkpoint with the lowest loss on the development set for evaluation. The experiment was conducted three times with different random seeds.

\subsection{Targeted syntactic evaluation}
In order to evaluate whether LMs learn human-like syntactic generalization, we employed six test circuits \citep{hu-etal-2020-systematic} on the SyntaxGym benchmark \citep{gauthier-etal-2020-syntaxgym}. Specifically, each test circuit deals with the following grammatical phenomenon: Agreement, Licensing, Garden-Path Effects, Gross Syntactic State, Center Embedding, and Long-Distance Dependencies. Each circuit is further subcategorized into suites; for example, the Agreement circuit contains a suite on a specific type of Agreement, such as ``subject-verb number agreement with prepositional phrase''. Each test suite consists of items designed to probe the specific grammatical phenomenon, and LMs succeed when they meet a success criterion, which defines inequalities among conditional probabilities on a grammatically critical position that should hold if they have learned the appropriate syntactic generalization. For example, to succeed on an item of the  ``subject-verb number agreement with prepositional phrase'' suite, LMs should assign a higher probability to the underlined critical position of (\ex{1}a) than (\ex{1}b):
\eenumsentence[1]{
\item \ac The author next to the senators \underline{is} good.
\item \un The author next to the senators \underline{are} good.
}

Following \citet{qian-etal-2021-structural}, we employed word-synchronous beam search \citep{stern-etal-2017-effective} to derive the probability of a grammatically critical position from syntactic LMs. Word-synchronous beam search retains a collection of the most likely syntactic structures that are predicted given an observed partial sentence $w_1, \cdots, w_{i}$ and marginalizes their probabilities to approximate $p(w_{i}|w_{<i})$:
\begin{align}
    \nonumber
    p(w_{i}|w_{<i}) &= \frac{p(w_1, \cdots, w_{i})}{p(w_1, \cdots, w_{i-1})}\\
    &\sim \frac{\sum_{Y_{i} \in \mathcal{Y}_{i}}{p(w_1, \cdots, w_{i}, Y_{i})}}{\sum_{Y_{i-1} \in \mathcal{Y}_{i-1}}{p(w_1, \cdots, w_{i-1}, Y_{i-1})}}
\end{align}
where $\mathcal{Y}_{i}$ denotes the collection of syntactic structures given $w_1, \cdots, w_{i}$. Following \citet{qian-etal-2021-structural}, we set the action beam size to 100, word beam size to 10, and fast-track to 5. 

\section{Results and discussion}
\subsection{Overall accuracies}
\label{subsec:overall}
\begin{figure}[t]
    \centering
    \includegraphics[width=\hsize]{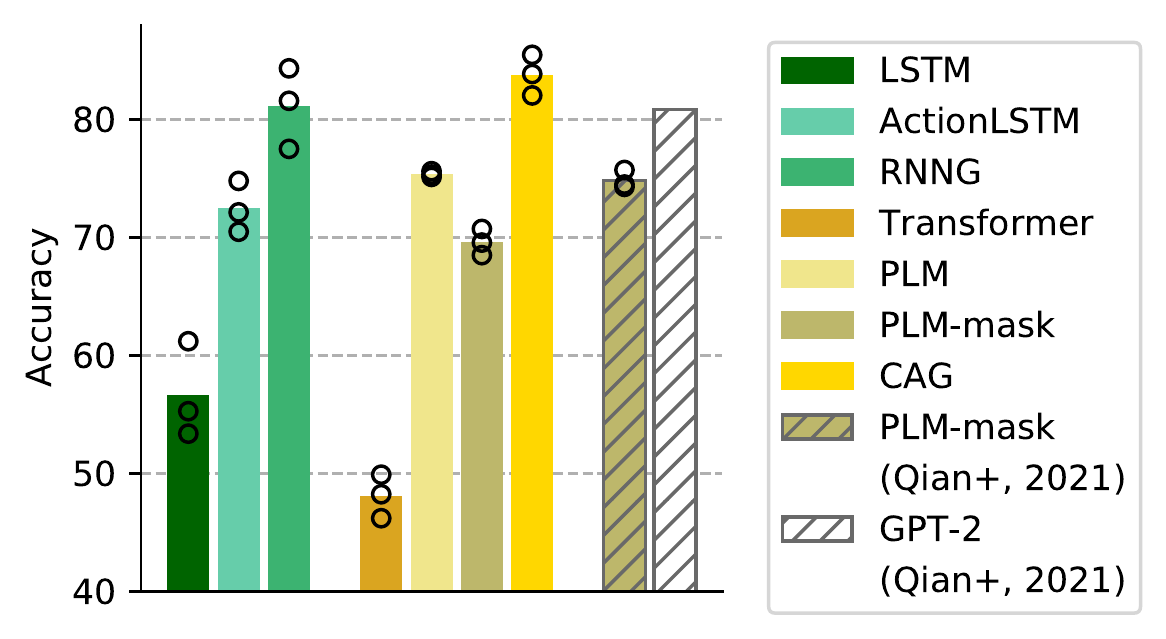}
    \caption{Overall accuracies of our controlled experiment. The average accuracies across the SyntaxGym test suites and different random seeds (the vertical axis) are plotted against the LMs investigated in this paper (the horizontal axis), with the accuracies of PLM-masks and GPT-2 from \citet{qian-etal-2021-structural}. The accuracies of PLM-masks and GPT-2 from \citet{qian-etal-2021-structural} are reference points as their model sizes are significantly larger than the other models investigated in this paper. Each dot denotes the accuracy of a specific seed.}
    \label{fig:overall}
\end{figure}

\begin{table*}[t]
    \centering
    \scalebox{0.66}{
    \begin{NiceTabular}{l||c|c|c|cc}
    \toprule
        \multicolumn{1}{c}{}&\multicolumn{1}{c}{$-\syn$}&\multicolumn{2}{c}{$+\syn$} & \multicolumn{2}{c}{}\\
        \cmidrule{3-4}
        \multicolumn{1}{c}{}&\multicolumn{1}{c}{}&\multicolumn{1}{c}{$-\comp$}&\multicolumn{1}{c|}{$+\comp$}&\multicolumn{1}{c}{$[+\sy]-[-\sy]$}&\multicolumn{1}{c}{$[+\com]-[-\com]$}\\
        \hline\hline
        $-\selfattn$ & 56.6 $\pm$ 3.3 (LSTM) & 72.5 $\pm$ 1.8 (ActionLSTM) & 81.1 $\pm$ 2.8 (RNNG) & 20.2 $\pm$ 4.7 & 8.6 $\pm$ 3.3\\
        \hline
        $+\selfattn$ & 48.1 $\pm$ 1.5 (Transformer) & 75.4 $\pm$ 0.2 (PLM)  & \begin{tabular}{c}
        69.6 $\pm$ 0.9 (PLM-mask) \\
        \textbf{83.8 $\pm$ 1.4 (CAG)}
        \end{tabular} & \begin{tabular}{c}
        24.4 $\pm$ 1.8 \\
        31.5 $\pm$ 2.1
        \end{tabular}  & \begin{tabular}{c}
        -5.8 $\pm$ 0.9 \\
        8.4 $\pm$ 1.4
        \end{tabular}\\
        \hline
        $[+\sa] - [-\sa]$ & -8.5 $\pm$ 3.7 & 2.9 $\pm$ 1.8 & \begin{tabular}{c}
        -11.5 $\pm$ 2.9 \\
        2.7 $\pm$ 3.1
        \end{tabular} & & \\
        \bottomrule 
    \end{NiceTabular}}
    \caption{Overall accuracy of each LM and the difference in the accuracy between minimally different LMs. $[+\sa] - [-\sa]$ denotes the difference in the accuracy between LMs with $+\selfattn$ and $-\selfattn$. $[+\sy]-[-\sy]$ and $[+\com]-[-\com]$ denote the differences in the accuracy between LMs with $+\syn$ and $-\syn$, and between LMs with $+\comp$ and $-\comp$, respectively. The standard deviations of the differences were calculated, assuming that the accuracies were normally distributed.}
    \label{tab:overall}
\end{table*}

Overall accuracies of our controlled experiment are summarized in Figure~\ref{fig:overall}. The average accuracies across the SyntaxGym test suites and different random seeds (the vertical axis) are plotted against the LMs investigated in this paper (the horizontal axis), with the accuracies of PLM-masks and GPT-2 \citep{radford2019language} from \citet{qian-etal-2021-structural}. The accuracies of PLM-masks and GPT-2 from \citet{qian-etal-2021-structural} are reference points as their model sizes are significantly larger than the other models investigated in this paper. Each dot denotes the accuracy of a specific seed. The results demonstrate that CAGs achieved the highest overall accuracy, suggesting that the composition function and the self-attention mechanism both play an important role to make LMs more human-like. Notice importantly that CAGs (83.8\%) outperformed GPT-2 (80.8\%) trained on 250$\times$ data with a 7$\times$ model size.

In the rest of this subsection, we discuss the effects of model components on the overall accuracy. In order to isolate the effects of individual components, Table~\ref{tab:overall} shows the overall accuracy of each LM and the difference in the accuracy between minimally different LMs.

\paragraph{$+$Syntax vs. $-$Syntax}
The LMs with explicit syntactic supervision outperformed the LMs without it, both without the self-attention mechanism (LSTM: 56.6\%, the average accuracy of ActionLSTM and RNNG: 76.7\%; +20.2\%) and with the self-attention mechanism (Transformer: 48.1\%, the average accuracy of PLM and PLM-mask: 72.5\%; +24.4\%, and the average accuracy of PLM and CAG: 79.4\%; +31.5\%). This result corroborates previous work \citep{kuncoro-etal-2017-recurrent, wilcox-etal-2019-structural, hu-etal-2020-systematic}, suggesting that explicit syntactic supervision plays an important role to make LMs more human-like.

\paragraph{$+$Composition vs. $-$Composition}
The LMs with the composition function outperformed the LMs without it, both without the self-attention mechanism (ActionLSTM: 72.5\%, RNNG: 81.1\%; +8.6\%) and with the self-attention mechanism (PLM: 75.4\%, CAG: 83.8\%; +8.4\%), suggesting that the composition function induces human-like syntactic generalization \citep{kuncoro-etal-2017-recurrent, wilcox-etal-2019-structural}.

\paragraph{$+$SelfAttn vs. $-$SelfAttn}
Without explicit syntactic supervision, the LMs with the self-attention mechanism underperformed the LMs without it (LSTM: 56.6\%, Transformer: 48.1\%; -8.5\%). 
In contrast, with explicit syntactic supervision, the LMs with the self-attention mechanism outperformed the LMs without it, both without the composition function (ActionLSTM: 72.5\%, PLM: 75.4\%; +2.9\%) and with the composition function (RNNG: 81.1\%, CAG: 83.8\%; +2.7\%). This result suggests that it is important to selectively attend to previous structural information not just words.

\begin{figure*}[t]
    \centering
    \includegraphics[width=\hsize]{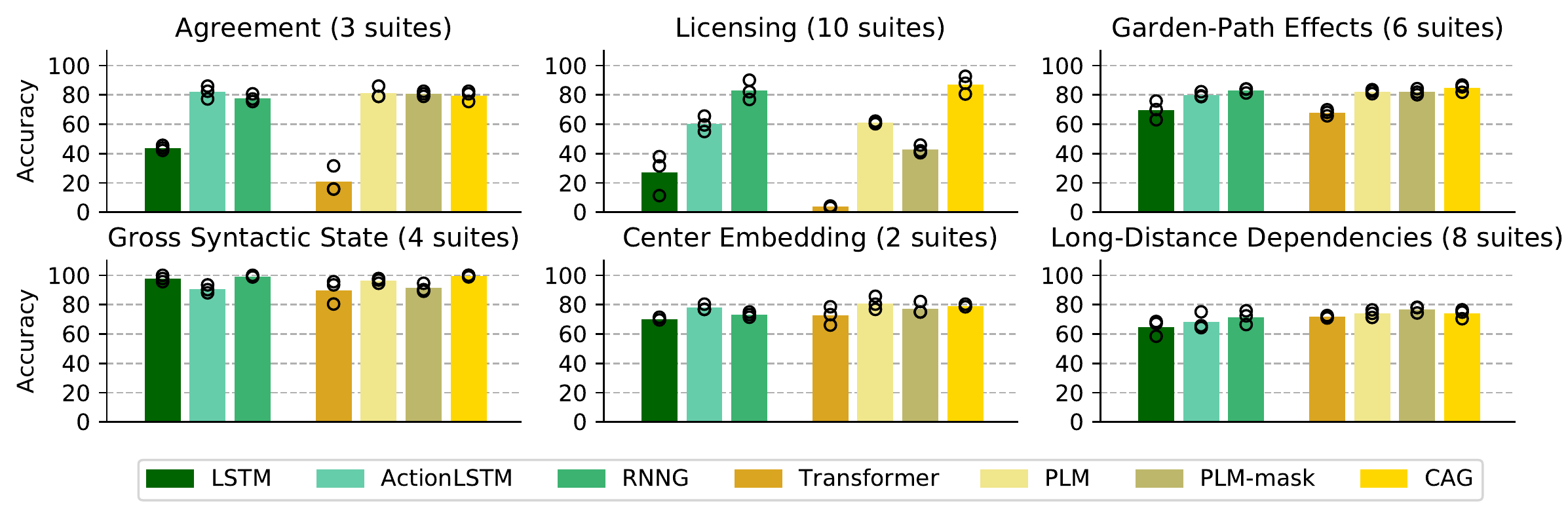}
    \caption{Circuit accuracies of our controlled experiment. The average accuracies across the SyntaxGym test suites and different random seeds on each test circuit (the vertical axis) are plotted against the LMs investigated in this paper (the horizontal axis). Each dot denotes the accuracy of a specific seed.}
    \label{fig:circuit}
\end{figure*}

\paragraph{$+$Composition vs. $(+)$Composition}
CAGs with the composition function outperformed PLM-masks with the dynamic masking mechanism (PLM-mask: 69.6\%, CAG: 83.8\%; +14.2\%). This result suggests that recursive composition of subtrees has additional advantages over the local subtree information in inducing human-like syntactic generalization. Note incidentally that our PLM-masks achieved a lower accuracy (69.6\%) than PLM-masks from \citet{qian-etal-2021-structural} (74.8\%), which may be caused by the difference in balance between specialized and vanilla attention heads: we specialized two out of 4 attention heads, whereas \citet{qian-etal-2021-structural} specialized two out of 12. Nevertheless, given that CAGs (83.8\%) outperformed the PLM-masks from \citet{qian-etal-2021-structural} (74.8\%) by a large margin, it is safe to conclude that recursive composition of subtrees has additional advantages over the local subtree information.

\subsection{Circuit accuracies}
Circuit accuracies of our controlled experiment are summarized in Figure~\ref{fig:circuit}. The average accuracies across the SyntaxGym test suites and different random seeds on each test circuit (the vertical axis) are plotted against the LMs investigated in this paper (the horizontal axis). Each dot denotes the accuracy of a specific seed. The results demonstrate that with explicit syntactic supervision, the LMs with the self-attention mechanism marginally outperformed the LMs without it on most of the test circuits, but the LMs with the composition function outperformed or underperformed the LMs without it depending on the test circuits. 

In the rest of this subsection, we investigate the pros and cons of the composition function through closer inspection of grammatical phenomena.

\paragraph{Syntactic features may percolate into the subtree representations.}
The LMs with the composition function outperformed the comparable LMs without it on three out of six circuits (Licensing, Garden-Path Effects, and Gross Syntactic State). Specifically, RNNGs and CAGs both outperformed ActionLSTMs and PLMs by a large margin (+23.0\% and +26.0\%, respectively) on Licensing, which includes items like (\ex{2}):
\eenumsentence[2]{
    \item \ac The author next to the senators hurt \\\ac \underline{herself}.
    \item \un The authors next to the senator hurt \\\ac \underline{herself}.
}
To successfully assign a higher probability to (\ex{2}a) than (\ex{2}b), LMs should understand that the reflexive pronoun must agree with the subject of the sentence in number. The subject NP ``The author/authors next to the senators/senator'' is composed into a single NP vector, as confirmed by the fact that RNNGs and CAGs both correctly assigned the following structure ``(NP The author/authors (ADVP next (PP to (NP the senators/senator))))'' to the subject NP.\footnote{RNNGs and CAGs both achieved considerably high bracketing F1 (RNNG: 96.0, CAG: 98.2) against acceptable test sentences parsed with the state-of-the-art constituency parser \citep{kitaev-klein-2018-constituency} on the Licensing circuit.} Given that RNNGs and CAGs successfully assigned a higher probability to an acceptable sentence through this subject NP vector, we can hypothesize that the syntactic features such as number may properly percolate into the subject NP vector.



\paragraph{Semantic features may not percolate into the subtree representations.}
In contrast, the LMs with the composition function underperformed the comparable LMs without it on the other circuits (Agreement, Center Embedding, and Long-Distance Dependencies). Specifically, RNNGs and CAGs both underperformed ActionLSTMs and PLMs most significantly on Center Embedding (-4.76\% and -1.79\%, respectively), which includes items like (\ex{3}):
\eenumsentence[3]{
    \item \ac The shirt that the man \underline{bought ripped}.
    \item \un The shirt that the man \underline{ripped bought}.
}
To successfully assign a higher probability to (\ex{3}a) than (\ex{3}b), LMs should understand that the verb that can take the inanimate subject ``shirt'' should appear at the end of the sentence. The subject NP ``The shirt that the man bought/ripped'' is composed into a single NP vector, as confirmed by the fact that RNNGs and CAGs both correctly assigned the following structure ``(NP The shirt (SBAR (WHNP that)(S (NP the man)(VP bought/ripped))))'' to the subject NP.\footnote{RNNGs and CAGs both achieved high bracketing F1 (RNNG: 96.7, CAG: 95.2) on the Center Embedding circuit. In addition, these scores are higher than ActionLSTMs and PLMs (ActionLSTM: 96.1, PLM: 94.2), respectively, indicating that the lower accuracy of RNNGs and CAGs than ActionLSTMs and PLMs on this circuit is not due to failure in parsing.} Given that RNNGs and CAGs failed to assign a higher probability to an acceptable sentence through this subject NP vector, we can hypothesize that the semantic features such as animacy may not properly percolate into the subject NP vector.



\paragraph{What kind of features percolates?}
The important implication here is that, with the composition function, the syntactic features may percolate into the subtree representations, but the semantic features may not. The detailed analysis of this implication (e.g., an analysis of the inner mechanics of feature percolation at the single neuron level; \citealp{lakretz-etal-2019-emergence}) will remain for future work.



\subsection{Overall accuracy and perplexity}
\begin{figure}[t]
    \centering
    \includegraphics[width=\hsize]{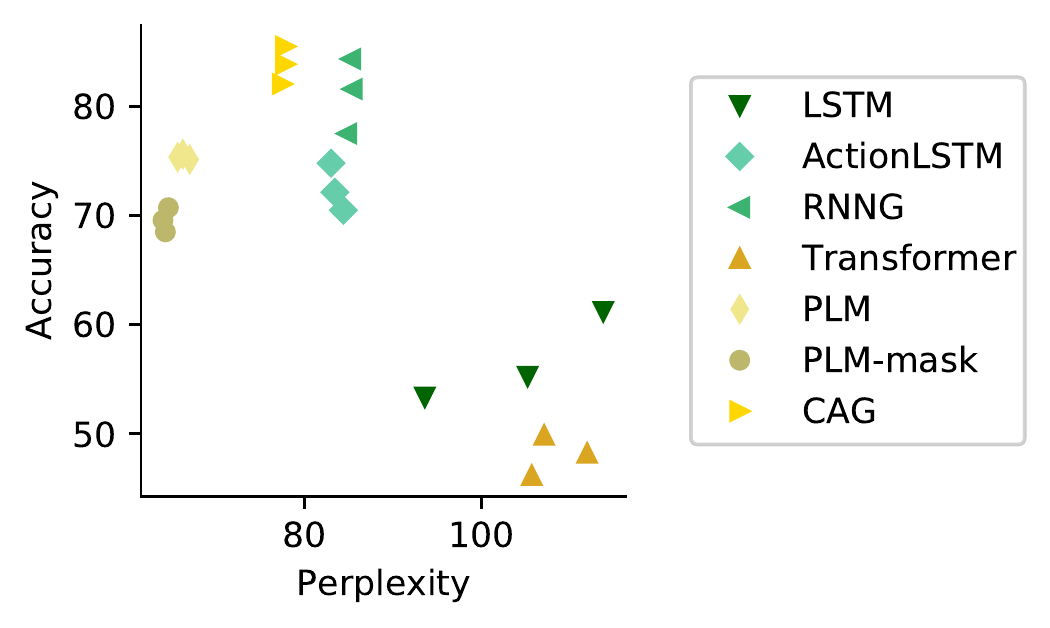}
    \caption{The relationship between the overall accuracy and perplexity: the overall accuracy (vertical axis) is plotted against perplexity (horizontal axis; lower is better).}
    \label{fig:ppl_acc}
\end{figure}

In this subsection, we compare the SyntaxGym overall accuracy against perplexity, the standard evaluation metric for LMs. The relationship between the overall accuracy and perplexity is summarized in Figure~\ref{fig:ppl_acc}: the overall accuracy (vertical axis) is plotted against perplexity (horizontal axis; lower is better). Following \citet{qian-etal-2021-structural}, we calculated the perplexity on the BLLIP held-out test set and derived the perplexity from the syntactic LMs, given the syntactic structures of the test sentences equal to the gold structures. Figure~\ref{fig:ppl_acc} demonstrates that explicit syntactic supervision generally improves both the overall accuracy and perplexity, but among the syntactic LMs, the overall accuracy is not linearly correlated with perplexity: PLMs and PLM-masks achieved worse overall accuracy, but better perplexity than RNNGs and CAGs. This result corroborates \citet{hu-etal-2020-systematic} that suggests a dissociation between perplexity and human-like syntactic generalization performance.

Recently, the relationship between perplexity and LMs' cognitive plausibility has attracted considerable attention. Besides LMs' human-like syntactic generalization performance, previous work on the correlation between perplexity and LMs' psychometric predictive power has typically reported that LMs with the better perplexity are more cognitively plausible \citep{fossum-levy-2012-sequential, goodkind-bicknell-2018-predictive, DBLP:conf/cogsci/WilcoxGHQL20}, but more recently, the counter-argument that lower perplexity is not always human-like has been widely discussed \citep{hao-etal-2020-probabilistic, oh-etal-2021-surprisal, kuribayashi-etal-2021-lower}. Given these recent trends, it is possible that the evaluation solely on perplexity may be orthogonal to the goal of human-like LMs (cf. \citealp{linzen-2020-accelerate}).

\section{Related work}
\label{sec:releted}
While writing this paper, we noticed that \citet{https://doi.org/10.48550/arxiv.2203.00633}, which is similar in spirit to our work, was submitted to the arXiv: they proposed Transformer Grammars (TGs) that incorporate recursive syntactic composition. TGs obtain a single vector representation of subtrees with the self-attention mechanism via an attention mask, but in contrast, CAGs obtain the representation with the composition function based on bidirectional LSTMs. While TGs are superior to CAGs in computational efficiency (see \hyperref[sec:lim]{Limitations} section), CAGs achieved better syntactic generalization performance on SyntaxGym (83.8\%) than TGs (82.5\%) that were trained with a 12$\times$ model size, suggesting that the composition function based on bidirectional LSTMs is advantageous in obtaining a vector representation of subtrees. Thorough comparisons between CAGs and TGs will remain for future work.


\section{Conclusion}
In this paper, we proposed a novel architecture called \textbf{Composition Attention Grammars} (CAGs) that recursively compose subtrees into a single vector representation with the composition function, and selectively attend to previous structural information with the self-attention mechanism. We investigated whether these components---the composition function and the self-attention mechanism---can both induce human-like syntactic generalization. Specifically, we trained LMs with and without these two components with the model sizes carefully controlled, and evaluated their syntactic generalization performance against six test circuits on the SyntaxGym benchmark. The results demonstrated that the composition function and the self-attention mechanism both play an important role to make LMs more human-like, and closer inspection of grammatical phenomena implied that the composition function allowed syntactic features, but not semantic features, to percolate into subtree representations.


\section*{Limitations}
\label{sec:lim}
Although it is not a central research question in this paper, a limitation with CAGs is their computational cost. While TGs \cite{https://doi.org/10.48550/arxiv.2203.00633} process all inputs simultaneously during training as in vanilla Transformers, CAGs must be trained recursively because the internal state of the stack changes dynamically due to the composition function. In fact, although we utilized effective batching for LMs with the composition function \citep{noji-oseki-2021-effective} and prevented CAGs from re-computing pre-computed attention keys and values, training of CAGs on the BLLIP-\textsc{lg} dataset (1.8M sentences and 42M tokens) for 15 epochs took two weeks on eight GPUs (NVIDIA V100). In addition, the self-attention mechanism consumes a large amount of memory, making it difficult to train CAGs with larger model sizes. The model size in this paper is the maximum that can be trained on V100 with 32GB memory. In order to address these limitations, we plan to introduce a computationally efficient self-attention mechanism (cf. \citealp{https://doi.org/10.48550/arxiv.2009.06732}) to CAGs in future work.

\section*{Acknowledgement}
\label{sec:ack}
We would like to thank Peng Qian for sharing the re-parsed BLLIP-\textsc{lg} dataset, which is used to train syntactic LMs in \citet{hu-etal-2020-systematic}, and for answering various questions. We are also grateful to three anonymous reviewers for valuable comments and suggestions. This work was supported by JST PRESTO Grant Number JPMJPR21C2.

\bibliography{emnlp2022}
\bibliographystyle{acl_natbib}

\clearpage
\appendix
\section{Effect of individual components on circuit accuracies.}
\label{app:circuit}
As with the overall accuracy, in order to isolate the effect of individual components on circuit accuracies, Table~\ref{tab:circuit} shows the circuit accuracy of each LM and the difference in accuracy between LMs with minimal differences.

\begin{table*}[t]
    \centering
    \begin{tabular}{c}
        \vspace{0.5cm}
        \begin{minipage}[t]{0.95\hsize}
            \centering
            \scalebox{0.63}{
            \begin{NiceTabular}{l|c|c|c||cc}
            \toprule
            \multicolumn{1}{c}{}&\multicolumn{1}{c}{$-\syn$}&\multicolumn{2}{c}{$+\syn$} & \multicolumn{2}{c}{}\\
            \cmidrule{3-4}
            \multicolumn{1}{c}{}&\multicolumn{1}{c}{}&\multicolumn{1}{c}{$-\comp$}&\multicolumn{1}{c|}{$+\comp$}&\multicolumn{1}{c}{$[+\sy]-[-\sy]$}&\multicolumn{1}{c}{$[+\com]-[-\com]$}\\
            \hline
            $-\selfattn$ & 43.9 $\pm$ 1.4 (LSTM) & \textbf{81.9 $\pm$ 3.6 (ActionLSTM)} & 77.8 $\pm$ 2.2 (RNNG) & 38.0 $\pm$ 3.9 & -4.09 $\pm$ 4.2\\
            \hline
            $+\selfattn$ & 21.1 $\pm$ 7.4 (Transformer) & 81.3 $\pm$ 3.3 (PLM)  & \begin{tabular}{c}
            80.7 $\pm$ 1.4 (PLM-mask) \\
            79.5 $\pm$ 3.0 (CAG)
            \end{tabular} & \begin{tabular}{c}
            59.9 $\pm$ 8.2\\
            59.3 $\pm$ 8.6
            \end{tabular} & \begin{tabular}{c}
            -0.585 $\pm$ 3.6 \\
            -1.75 $\pm$ 4.5
            \end{tabular}\\
            \hline
            \hline
            $[+\sa] - [-\sa]$ & -22.8 $\pm$ 7.6 & -0.585 $\pm$ 4.9 & \begin{tabular}{c}
            2.92 $\pm$ 2.6 \\
            1.75 $\pm$ 3.7
            \end{tabular} & & \\
            \bottomrule  
            \end{NiceTabular}}
            \subcaption{Agreement}
            \label{tbl:agreement}
        \end{minipage}\\
        \vspace{0.5cm}
        \begin{minipage}[t]{0.95\hsize}
            \centering
            \scalebox{0.63}{
            \begin{NiceTabular}{l|c|c|c||cc}
            \toprule
            \multicolumn{1}{c}{}&\multicolumn{1}{c}{$-\syn$}&\multicolumn{2}{c}{$+\syn$} & \multicolumn{2}{c}{}\\
            \cmidrule{3-4}
            \multicolumn{1}{c}{}&\multicolumn{1}{c}{}&\multicolumn{1}{c}{$-\comp$}&\multicolumn{1}{c|}{$+\comp$}&\multicolumn{1}{c}{$[+\sy]-[-\sy]$}&\multicolumn{1}{c}{$[+\com]-[-\com]$}\\
            \hline
            $-\selfattn$ & 26.9 $\pm$ 11.3 (LSTM) & 60.0 $\pm$ 4.3 (ActionLSTM) & 83.0 $\pm$ 5.4 (RNNG) & 33.1 $\pm$ 12 & 23.0 $\pm$ 6.9\\
            \hline
            $+\selfattn$ & 3.68 $\pm$ 0.4 (Transformer) & 61.1 $\pm$ 0.8 (PLM)  & \begin{tabular}{c}
            42.7 $\pm$ 2.2 (PLM-mask) \\
            \textbf{87.0 $\pm$ 5.0 (CAG)}
            \end{tabular} & \begin{tabular}{c}
            48.2 $\pm$ 2.4 \\
            70.4 $\pm$ 5.1
            \end{tabular} & \begin{tabular}{c}
            -18.3 $\pm$ 2.4 \\
            26.0 $\pm$ 5.0
            \end{tabular}\\
            \hline
            \hline
            $[+\sa] - [-\sa]$ & -23.2 $\pm$ 11 & 1.05 $\pm$ 4.4 & \begin{tabular}{c}
            -40.3 $\pm$ 5.9 \\
            4.04 $\pm$ 7.4
            \end{tabular} & & \\
            \bottomrule  
            \end{NiceTabular}}
            \subcaption{Licensing}
            \label{tbl:licensing}
        \end{minipage}\\
        \vspace{0.5cm}
        \begin{minipage}[t]{0.95\hsize}
            \centering
            \scalebox{0.63}{
            \begin{NiceTabular}{l|c|c|c||cc}
            \toprule
            \multicolumn{1}{c}{}&\multicolumn{1}{c}{$-\syn$}&\multicolumn{2}{c}{$+\syn$} & \multicolumn{2}{c}{}\\
            \cmidrule{3-4}
            \multicolumn{1}{c}{}&\multicolumn{1}{c}{}&\multicolumn{1}{c}{$-\comp$}&\multicolumn{1}{c|}{$+\comp$}&\multicolumn{1}{c}{$[+\sy]-[-\sy]$}&\multicolumn{1}{c}{$[+\com]-[-\com]$}\\
            \hline
            $-\selfattn$ & 69.6 $\pm$ 5.2 (LSTM) & 80.1 $\pm$ 1.5 (ActionLSTM) & 83.1 $\pm$ 1.3 (RNNG) & 10.5 $\pm$ 5.4 & 3.01 $\pm$ 2.0\\
            \hline
            $+\selfattn$ & 67.9 $\pm$ 1.7 (Transformer) & 82.2 $\pm$ 1.1 (PLM)  & \begin{tabular}{c}
            82.0 $\pm$ 1.7 (PLM-mask) \\
            \textbf{84.6 $\pm$ 2.1 (CAG)}
            \end{tabular} & \begin{tabular}{c}
            14.2 $\pm$ 2.6 \\
            15.5 $\pm$ 2.9
            \end{tabular} & \begin{tabular}{c}
            -0.198 $\pm$ 2.1 \\
            2.38 $\pm$ 2.4
            \end{tabular}\\
            \hline
            \hline
            $[+\sa] - [-\sa]$ & -1.72 $\pm$ 5.5 & 2.18 $\pm$ 1.9 & \begin{tabular}{c}
            -1.03 $\pm$ 2.1 \\
            1.55 $\pm$ 2.5
            \end{tabular} & & \\
            \bottomrule  
            \end{NiceTabular}}
            \subcaption{Garden-Path Effects}
            \label{tbl:garden-path}
        \end{minipage}\\
        \vspace{0.5cm}
        \begin{minipage}[t]{0.95\hsize}
            \centering
            \scalebox{0.63}{
            \begin{NiceTabular}{l|c|c|c||cc}
            \toprule
            \multicolumn{1}{c}{}&\multicolumn{1}{c}{$-\syn$}&\multicolumn{2}{c}{$+\syn$} & \multicolumn{2}{c}{}\\
            \cmidrule{3-4}
            \multicolumn{1}{c}{}&\multicolumn{1}{c}{}&\multicolumn{1}{c}{$-\comp$}&\multicolumn{1}{c|}{$+\comp$}&\multicolumn{1}{c}{$[+\sy]-[-\sy]$}&\multicolumn{1}{c}{$[+\com]-[-\com]$}\\
            \hline
            $-\selfattn$ & 97.8 $\pm$ 1.8 (LSTM) & 90.6 $\pm$ 2.2 (ActionLSTM) & 99.3 $\pm$ 0.5 (RNNG) & -7.25 $\pm$ 2.9 & 8.70 $\pm$ 2.3\\
            \hline
            $+\selfattn$ & 89.9 $\pm$ 6.7 (Transformer) & 96.4 $\pm$ 1.4 (PLM)  & \begin{tabular}{c}
            91.3 $\pm$ 2.3 (PLM-mask) \\
            \textbf{99.6 $\pm$ 0.5 (CAG)}
            \end{tabular} & \begin{tabular}{c}
            3.95 $\pm$ 7.2 \\
            8.1 $\pm$ 6.9
            \end{tabular}  & \begin{tabular}{c}
            -5.07 $\pm$ 2.7 \\
            3.26 $\pm$ 1.4
            \end{tabular}\\
            \hline
            \hline
            $[+\sa] - [-\sa]$ & -7.97 $\pm$ 7.0 & 5.80 $\pm$ 2.6 & \begin{tabular}{c}
            -7.97 $\pm$ 2.4 \\
            0.362 $\pm$ 0.72
            \end{tabular} & & \\
            \bottomrule  
            \end{NiceTabular}}
            \subcaption{Gross Syntactic State}
            \label{tbl:gross_syntactic_state}
        \end{minipage}\\
        \vspace{0.5cm}
        \begin{minipage}[t]{0.95\hsize}
            \centering
            \scalebox{0.63}{
            \begin{NiceTabular}{l|c|c|c||cc}
            \toprule
            \multicolumn{1}{c}{}&\multicolumn{1}{c}{$-\syn$}&\multicolumn{2}{c}{$+\syn$} & \multicolumn{2}{c}{}\\
            \cmidrule{3-4}
            \multicolumn{1}{c}{}&\multicolumn{1}{c}{}&\multicolumn{1}{c}{$-\comp$}&\multicolumn{1}{c|}{$+\comp$}&\multicolumn{1}{c}{$[+\sy]-[-\sy]$}&\multicolumn{1}{c}{$[+\com]-[-\com]$}\\
            \hline
            $-\selfattn$ & 70.2 $\pm$ 0.8 (LSTM) & 78.0 $\pm$ 1.7 (ActionLSTM) & 73.2 $\pm$ 1.5 (RNNG) & 7.74 $\pm$ 1.9 & -4.76 $\pm$ 2.2\\
            \hline
            $+\selfattn$ & 72.6 $\pm$ 5.1 (Transformer) & \textbf{81.0 $\pm$ 3.7 (PLM)}  & \begin{tabular}{c}
            77.4 $\pm$ 3.4 (PLM-mask) \\
            79.2 $\pm$ 0.8 (CAG)
            \end{tabular} & \begin{tabular}{c}
            6.6 $\pm$ 7.2 \\
            7.5 $\pm$ 6.4
            \end{tabular}  & \begin{tabular}{c}
            -3.57 $\pm$ 5.0 \\
            -1.79 $\pm$ 3.8
            \end{tabular}\\
            \hline
            \hline
            $[+\sa] - [-\sa]$ & 2.38 $\pm$ 5.2 & 2.98 $\pm$ 4.0 & \begin{tabular}{c}
            4.17 $\pm$ 3.7 \\
            5.95 $\pm$ 1.7
            \end{tabular} & & \\
            \bottomrule  
            \end{NiceTabular}}
            \subcaption{Center Embedding}
            \label{tbl:center_emb}
        \end{minipage}\\
        \begin{minipage}[t]{0.95\hsize}
            \centering
            \scalebox{0.63}{
            \begin{NiceTabular}{l|c|c|c||cc}
            \toprule
            \multicolumn{1}{c}{}&\multicolumn{1}{c}{$-\syn$}&\multicolumn{2}{c}{$+\syn$} & \multicolumn{2}{c}{}\\
            \cmidrule{3-4}
            \multicolumn{1}{c}{}&\multicolumn{1}{c}{}&\multicolumn{1}{c}{$-\comp$}&\multicolumn{1}{c|}{$+\comp$}&\multicolumn{1}{c}{$[+\sy]-[-\sy]$}&\multicolumn{1}{c}{$[+\com]-[-\com]$}\\
            \hline
            $-\selfattn$ & 64.7 $\pm$ 4.5 (LSTM) & 68.4 $\pm$ 4.8 (ActionLSTM) & 71.5 $\pm$ 3.9 (RNNG) & 3.63 $\pm$ 6.6 & 3.17 $\pm$ 6.2\\
            \hline
            $+\selfattn$ & 71.9 $\pm$ 0.7 (Transformer) & 73.9 $\pm$ 2.1 (PLM)  & \begin{tabular}{c}
            \textbf{76.9 $\pm$ 1.8 (PLM-mask)} \\
            73.9 $\pm$ 2.6 (CAG)
            \end{tabular} & \begin{tabular}{c}
            3.5 $\pm$ 2.9 \\
            2.0 $\pm$ 3.4
            \end{tabular}  & \begin{tabular}{c}
            2.93 $\pm$ 2.8 \\
            0.00992 $\pm$ 3.3
            \end{tabular}\\
            \hline
            \hline
            $[+\sa] - [-\sa]$ & 7.20 $\pm$ 4.5 & 5.56 $\pm$ 5.2 & \begin{tabular}{c}
            5.32 $\pm$ 4.3 \\
            2.40 $\pm$ 4.7
            \end{tabular} & & \\
            \bottomrule  
            \end{NiceTabular}}
            \subcaption{Long-Distance Dependencies}
            \label{tbl:long-distance dependency}
        \end{minipage}\\
    \end{tabular}
    \caption{Circuit accuracy of each LM and the difference in the accuracy between LMs with minimal differences. $[+\sa] - [-\sa]$ denotes the difference in the accuracy between LMs with $+\selfattn$ and $-\selfattn$. $[+\sy]-[-\sy]$ and $[+\com]-[-\com]$ denote the difference in the accuracy between LMs with $+\syn$ and $-\syn$ , and the difference in the accuracy between LMs with $+\comp$ and $-\comp$, respectively. The standard deviations of the differences were calculated assuming that the accuracies were normally distributed.}
    \label{tab:circuit}
\end{table*}

\end{document}